# Recent Standard Development Activities on Video Coding for Machines

Wen Gao, *Member, IEEE*, Shan Liu, *Senior Member, IEEE*, Xiaozhong Xu, *Member, IEEE,*
Manouchehr Rafie, *Member, IEEE,* Yuan Zhang, Igor Curcio, *Member, IEEE*

*Abstract*—In recent years, video data has dominated internet traffic and becomes one of the major data formats. With the emerging 5G and internet of things (IoT) technologies, more and more videos are generated by edge devices, sent across networks, and consumed by machines. The volume of video consumed by machine is exceeding the volume of video consumed by humans. Machine vision tasks include object detection, segmentation, tracking, and other machine-based applications, which are quite different from those for human consumption. On the other hand, due to large volumes of video data, it is essential to compress video before transmission. Thus, efficient video coding for machines (VCM) has become an important topic in academia and industry. In July 2019, the international standardization organization, i.e., MPEG, created an Ad-Hoc group named VCM to study the requirements for potential standardization work. In this paper, we will address the recent development activities in the MPEG VCM group. Specifically, we will first provide an overview of the MPEG VCM group including use cases, requirements, processing pipelines, plan for potential VCM standards, followed by the evaluation framework including machine-vision tasks, dataset, evaluation metrics, and anchor generation. We then introduce technology solutions proposed so far and discuss the recent responses to the Call for Evidence issued by MPEG VCM group.

*Index Terms*—Video codec, Video coding for machine

## I. INTRODUCTION

VIDEO, as one of the major media formats, has occupied a very large portion of internet traffic in recent years. Video data has become the largest source of data consumed globally. There is a growing awareness that the majority of video traffic will be used by machines. Today's societies are becoming ever more multimedia-centric, data-dependent, and automated. Automation, analysis, and intelligence is moving beyond humans to "machine-specific" applications, creating the need for machine-to-machine (M2M) or machine-to-human (M2H) communications. The rise of AI-driven video intelligent solutions, such as video coding for machine (VCM) standards [1] for M2M or M2H vision, will be key solutions addressing the most severe challenges of multimedia computing, transmission, and storage. VCM will be transforming everyday video content by identifying, classifying, and indexing objects that appear within, so that the metadata becomes machine-specific, searchable, and actionable. It is expected that the trend will continue due to the convergence of emerging technologies such as 5G, artificial intelligence (AI), smart sensors, the internet of things (IoT), and connected and autonomous vehicles. The switch to AI-enabled 5G networks is happening now and is aiming to transform smart cities, the automotive industry, and intelligent transportation systems (ITS). Additionally, more and more edge devices can capture video signals, which are sent across either internet or private networks and consumed by machines for analysis. The emerging VCM standard can help mainstream visual data applications to broaden their use cases in the areas of autonomous cars, smart cities, smart sensors, intelligent industry, immersive entertainment, and beyond [2]. In most of these use cases, certain portions of videos are mainly used for machine-vision tasks such as image classification, object detection, segmentation, tracking, or similar applications. In some other use cases such as surveillance, humans may occasionally inspect some of the videos to extract additional information that is not captured by machines. Due to the huge volume of video data, video coding technologies have been employed to compress videos before transmission or storage. Traditionally video is consumed by human beings for a variety of usages such as entertainment, education, etc. Thus, video coding often utilizes characteristics of the human visual system (HVS) for better compression efficiency while maintaining good subjective quality. For example, the popular video coding standards, such as MPEG2, H.264/AVC, H.265/HEVC, and the recently finalized H.266/Versatile Video Coding (VVC), all follow this design principle.

Machine vision tasks are different from human vision tasks with different purposes and evaluation metrics. How to encode video for machine consumption becomes an interesting and challenging problem. This leads to the creation of an Ad-Hoc group so-called Video Coding for Machines (VCM) in July 2019 under the international standard organization MPEG, which also developed the popular video coding standards mentioned above.

This paragraph of the first footnote will contain the date on which you submitted your paper for review. It will also contain support information, including sponsor and financial support acknowledgment.

Wen Gao, Shan Liu, and Xiaozhong Xu are with Tencent Media Lab, Tencent America, 2747 Park Blvd, Palo Alto, CA 94306 USA (e-mail: wengao@tencent.com, shanl@tencent.com, xiaozhongxu@tencent.com).

Manouchehr Rafie is with Gyrfalcon Technology Inc., 1900 McCarthy Blvd. Suite 412, Milpitas, CA 95036 USA. (e-mail: m.rafie@gyrfalcontech.com).

Yuan Zhang is with China Telecom Research Institute, China (e-mail: zhangy666@chinatelecom.cn).

Igor Curcio is with Nokia Technologies, Finland (e-mail: igor.curcio@nokia.com).

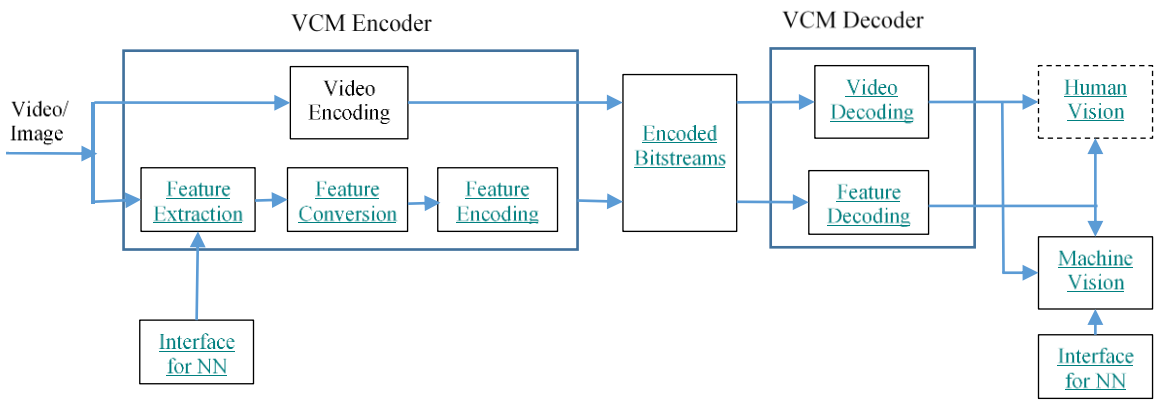

Fig 1. An example of potential VCM architecture [1]

The mandate of the MPEG VCM group can be summarized as follows: (1) Define use cases and related requirements for compression for machine vision and hybrid human/machine visions; (2) Collect dataset with ground truth and evaluation metrics; (3) Solicit technology evidence for feature compression, combined human/machine-oriented video representation and compression; (4) Develop a framework to evaluate and compare different technology solutions. (5) Develop the standards for video coding for machines.

The MPEG VCM group has attracted a lot of attention from experts in industry and academia and has made a lot of progress since its inception. In this paper, we will address the recent activities in the VCM group and would encourage more experts to join the study and standardization efforts for VCM.

This paper is organized as follows: We first introduce the related works and provide an overview of the VCM group including use cases, requirements, processing pipeline, and the timeline for potential VCM standards, followed by the evaluation framework adopted by the group to compare different technology solutions. We then discuss technology solutions submitted so far to the VCM group and the recent responses to the Call for Evidence (CfE) issued by the group. Discussions and conclusions are provided at the end.

## II. RELATED WORKS

In literature, [3] presents an overview on video coding for machine. Specifically, the authors reviewed traditional video coding for human vision and the standardization efforts on feature compression, i.e., MPEG-7 compact feature descriptor standards including Compact descriptor for Visual Search (CDVS) and Compact descriptor for Video Analysis (CDVA). A collaborative framework for joint video coding for human consumption and feature compression is introduced. Two potential solutions and the corresponding preliminary experimental results are presented and listed as follows:
- Solution 1: Deep intermediate feature compression for different tasks [4].
- Solution 2: Joint compression of feature and video.

The main difference between [3] and this paper is their focus. This paper will focus more on standardization efforts in MPEG-VCM group while [3] focuses more on a theoretic framework for video coding for machine. Furthermore, [3] intends to provide a paradigm for collaborative video and feature compression, which may be served as a potential solution satisfying the requirements of the VCM group. Actually Solution 2 mentioned in [3] has been proposed to MPEG-VCM as a preliminary evidence, which will be discussed later [55]. In addition, an overview of [3] was introduced to the VCM group in [57].

Intermediate feature compression is an important topic. A strategy to represent and transmit the deep learning based intermediate layer feature maps is introduced in [4], which enables a good balance among computation load at server end, transmission bandwidth, and generalization capability for large scale cloud-based machine analysis tasks. Generic data compression tools, i.e., GZIP, ZLIB, BZIP2 and LZMA, are used for lossless compression of the intermediate features. For lossy compression, multiple feature maps are treated as a video sequence and compressed using an HEVC encoder. Note that majority of feature compression solutions proposed to the MPEG-VCM group follow the similar idea and part of this work is also presented in MPEG-VCM as a potential evidence [33].

A semantically structured image coding (SSIC) framework for multiple machine vision tasks is addressed in [5]. In this framework, high-level features, such as class IDs and bounding boxes for objects in an image, and low-level features for all objects and background are extracted and compressed separately. The bitstreams for high-level features and low-level features are combined into a semantically structured bitstream (SSB). Depending on the machine vision tasks at the server side, the front-end device only needs to send part of the SSB, which is decoded to generate reconstructed features and/or reconstructed images as input to the machine vision tasks or directly consumed by humans at the server side. Experimental results show that this scheme can achieve performance comparable to existing image coding scheme. Note that part of the scheme in [5] is also introduced as a potential evidence to MPEG-VCM group [53].

## III. OVERVIEW OF MPEG VCM

The MPEG VCM group was established in July 2019 with the target to create an international standard for VCM to bring interoperability among edge devices and servers. Before the start of actual standardization work, it is necessary to collect use cases and requirements serving as guidelines for the coming VCM standardization work.

The MPEG VCM activity aims to standardize a bitstream format generated by compressing a previously extracted feature stream or video stream. Fig. 1 [1] shows an example of potential VCM architecture. The VCM codec could be a video codec or a feature codec, or both. In the case of feature codec, the VCM feature encoding consists of feature extraction, feature conversion/packing, and feature coding. There may be an interface to an external Neural Network Compression and Representation (NNR) for the feature extraction and the task-specific networks. Regarding the detailed VCM processing requirements and pipelines, please refer to the Evaluation Framework of Video Coding for Machines document [6].

Through a collaborative process, the group has created three possible processing pipelines covering different use cases and establish a tentative timeline for the initial study phase and the formal standardization phase. We will discuss each topic separately in this section.

*A. Use Cases*

The MPEG VCM group has identified six classes of use cases [1]. Among them, three classes are exemplary, shown as follows:

a) **Intelligent Transportation**
   To perform tasks such as object detection, semantic segmentation, lane tracking, traffic monitoring at the local (edge) level in real-time and taking driving actions accordingly, sensors in the infrastructure may communicate features to different vehicles, and/or vehicles may need to communicate features among each other. Video reconstruction for human consumptions should be possible when needed. In addition, non-visible light images such as IR (infrared) or LiDAR (light detection and ranging) images may be employed in some scenarios.

b) **Smart City**
   With the rise of the Internet of Things (IoT), there exists a high degree of interconnectivity between different nodes/devices, which often communicate with each other to share information. The typical smart city applications include traffic monitoring, density detection and prediction, traffic flow prediction, and resource allocation. Video surveillance use case is a subset of smart city, which often requires a huge amount of storage due to a large number of camera sensors and the long duration of videos to be recorded. In this case, the typical machine vision tasks are object detection, instance segmentation, and key point detection. It is also desirable to be able to reconstruct videos for human consumption.

c) **Intelligent Contents:**
   In recent years, a large number of images/videos are generated by professionals or by amateurs and shared on the internet. How to protect certain age groups (i.e., people under 18) from inappropriate contents is an important issue. The traditional manual review process is time-consuming and labor-intensive. Machine vision technologies can help review, rate, process, and distribute a variety of forms of images/videos such as live videos, short video clips, etc. The main machine vision tasks are similar to those in surveillance.

*B. Requirements*

Based on the above use cases, the VCM group is able to identify the key requirements for the potential VCM standards [1]. The mandatory requirements are listed as follows:

a) **Efficient compression performance**
   The size of coded bitstreams representing videos or features shall be less than that coded under traditional video coding techniques with comparable performance and at a reasonable setting.

b) **Ability to support one or more tasks**
   The generated coded bitstreams shall be usable and optimized for different scenarios: one bitstream for a single task, one common bitstream for multiple tasks.

c) **Varying degrees of performance for multiple tasks**
   The coding technology shall support varying levels of quality as measured by the performance of different tasks.

In addition, the group also specifies the following optional or recommended requirements. (1) Hybrid machine and human consumption: A common bitstream should be used for machine and human consumption, additional bitstream(s) is optional; (2) Computational offloading: The coding technology has the ability to offload part of machine vision tasks into front end devices, and the outputs of intermediate layers of neural networks can be compressed and transmitted to cloud servers; (3) Privacy protection: The coded bitstreams enable different levels of ability to reconstruct images/video according to the predefined privacy level.

*C. Processing Pipeline*

The VCM group focuses on the uses cases that require compression of a video or features extracted from a video. Three possible VCM processing pipelines are suggested and shown in Fig. 2 [6].

In Pipeline 1, video codecs are used to compress videos before transmission or storage and decompress the generated bitstreams into videos for consumption by machine analysis devices. There is no restriction on the architecture of the video codec, which can follow the traditional hybrid block-based codec or deep learning-based codec that optimized for machine vision or hybrid human and machine vision. Note that the anchor generation mentioned later follows this pipeline, where VVC is used as the codec.

In Pipeline 2, the neural network for a machine vision task is split into two parts: Part 1 network will be at edge devices that contain image sensors or video cameras. Part 2 network will be in servers that generate machine task inference results. There are two alternative approaches, one is to pack the output features of the Part 1 network into images or videos which are compressed by video encoders as shown in Pipeline 2a; the other is to directly encode the features according to their characteristics as shown in Pipeline 2b; the generated bitstream will be transmitted to servers through networks. At the server-side, video decoders decode the bitstream to generate reconstructed images/videos which are inversely packed into reconstructed features used as input to the Part 2 network shown in Pipeline 2a; In Pipeline 2b, feature decoders are used to decode the bitstreams into features that are fed into Part 2 network directly.

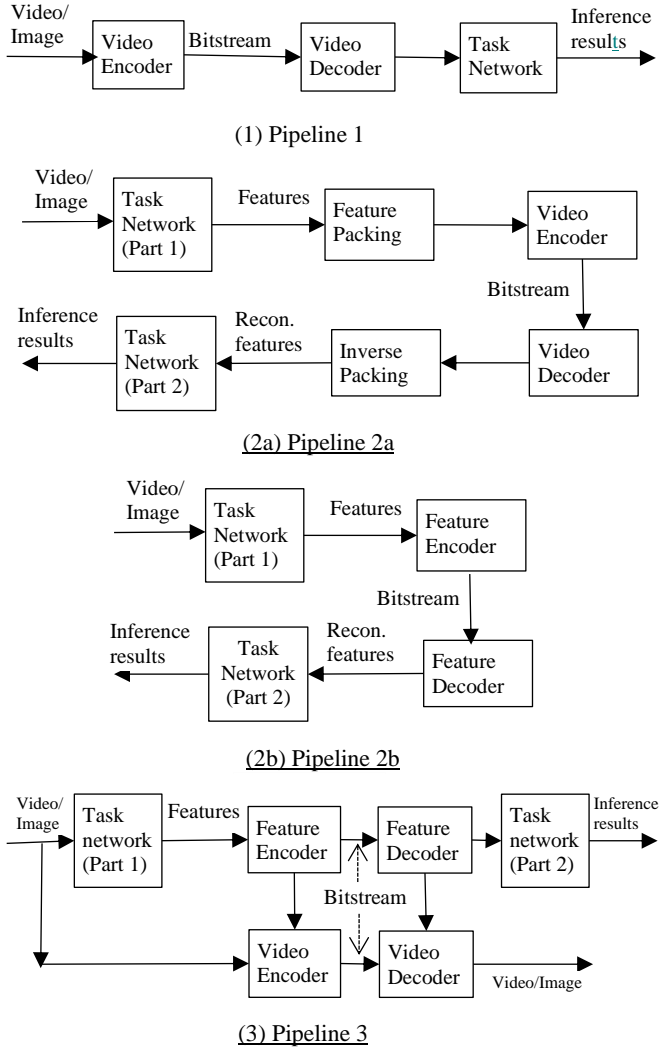

Fig 2. Proposed processing pipeline in VCM

Pipeline 3 handles the case of hybrid human and machine vision. The top branch of Pipeline 3 is a simplified version of Pipeline 2 that compresses the intermediate features while the bottom branch shows the encoding of the input video using a video encoder, which makes use of the raw video and the features extracted from the video as the inputs. This is different from the traditional video codec such as AVC, HEVC, or VVC. The video decoder at the bottom branch utilizes the generated bitstream and the reconstructed features from the top branch to reconstruct videos/images for human consumption.

One thing worth mentioning is that besides the above three pipelines, other pipelines are also possible and can be proposed to the group.

### D. Timeline

After its establishment in July 2019, the MPEG-VCM group has been soliciting technology evidences from the industry and academia and defined use cases, requirements and evaluation framework. To assess whether sufficient evidences exist for the standardization work, the group issued its first round of CfE in January 2021 [7]. The received responses to the CfE were evaluated in the MPEG meeting in April 2021. Based on the responses, the group concluded that meaningful evidences were presented for object detection tasks. However, evidences for other machine tasks such as object segmentation, object tracking, etc. and for multi-task user cases are still missing. It was decided to have another meeting cycle for further evidence collection. In the next MPEG meeting in July 2021, if the group decides the current technologies are mature enough, the formal standardization work may be launched. The standardization process usually takes about two to three years, depending on how the technologies evolve and the target set by the group. We will discuss the responses to the CfE in details in Section VI.

### IV. EVALUATION METHODOLOGIES

To compare different technical solutions, the MPEG VCM group has established an evaluation framework that includes machine vision tasks, evaluation dataset, evaluation metrics, and anchor generation [6].

### A. Machine Vision Tasks

There are many possible machine vision tasks in practice. From Section III, we can identify the following typical tasks: (1) Object detection (2) object (instance) segmentation (3) object tracking (4) Action Recognition, and (5) Pose Estimation. Note that both Action Recognition and Pose Estimation are built on top of key point detection mentioned in Section III. The details of the five machine vision tasks are provided in Table 1.

Table 1. information about machine vision tasks

| Task | Network Architecture | Link | Training Dataset |
|---|---|---|---|
| Object Detection | Faster R-CNN with ResNeXt-101 backbone | [8] | COCO train2017[12] |
| Object Segmentation | Mask R-CNN with ResNeXt-101 backbone | [8] | COCO train2017[12] |
| Object Tracking | JDE-1088x608 | [9] | HiEve training dataset[13] |
| Action Recognition | SlowFast | [10] | HiEve training dataset[13] |
| Pose Estimation | HRNet | [11] | COCO train2017[12] MPII Human Pose [14] |

In Table 1, for all machine vision tasks, the network architectures, the corresponding weblinks and training datasets used to obtain model parameters are specified. When evaluating a VCM solution, the task model parameters are expected to be fixed for fair comparison unless special reasons are provided. In addition, the model for object detection using the dataset FLIR [15] is retrained using the FLIR training dataset, which will be described in the following section.

*B. Evaluation datasets*

There are a variety of public datasets for machine learning, as summarized in [16]. However, a large portion of the datasets only permit non-commercial usage, which prohibits their usage for many participants in the VCM group. For some datasets, such as COCO [12], the annotation can be used with permissible license terms, but the underlying images/videos have restricted usage terms. Among those that allow for commercial usage, the characteristics may not be suitable to be used as evaluation datasets for VCM. Due to above reasons the group have identified the following five datasets for evaluation of VCM solutions, described as follows.

(1) OpenImageV6 [17]: OpenImageV6 is used for object detection and object segmentation. Since there are more than 20,000 images in its evaluation dataset, only 5000 of them are chosen to reduce the simulation time. Note that the set of 5000 images for object detection is not the same as the set for object segmentation [18].

(2) FLIR [15]: FLIR includes both RGB images and Infrared images that are suitable for object detection used in autonomous driving and advanced driver assistance scenarios (ADAS). Simulations show that the Infrared images achieve better detection performance than the RGB images at low light conditions. Thus, only the Infrared images are adopted for VCM evaluation [19]. Examples of Infrared images are shown in Fig. 3.

(3) HiEve-10 [20]: Hi-Eve dataset includes a large number of poses (>1M), complexity-event actions labels (>56K) and trajectories with long durations (with average trajectory length > 480 frames), which is suitable for evaluation of object tracking, action recognition and pose estimation. A subset of 10 videos in the HiEve dataset, called HiEve-10, can be used for commercial purpose. Within HiEve-10 dataset, the 7 training videos with annotations are chosen for VCM evaluation.

(4) TVD [21, 22] An open dataset was introduced to the group [21]. The dataset contains 86 clips. Each clip includes 65 frames with a resolution of 3840x2160. The contents of the sequences are quite diverse, ranging from a pedestrian walking on the street, bicycle riders, tea making, cooking, tourist site-seeing, and few others. In [22], An image dataset is obtained by sampling the TVD video sequences and all the images in the dataset are converted to 1920x1080 format. The corresponding bounding box annotations for all images and anchor results for object detection are provided as well.

(5) SFU-HW-Object-v1: Object detection annotations for 18 classes of object in video sequences used in HEVC standard development are provided in [23]. Anchor results for object detection using this dataset is provided in [24].

*C. Evaluation metric for single task*

Different machine vision tasks adopt different evaluation metrics, as shown in Table 2.

Note that in Table 2, the evaluation datasets are also included for each machine vision task since for object detection, the four datasets use slightly different evaluation metrics.

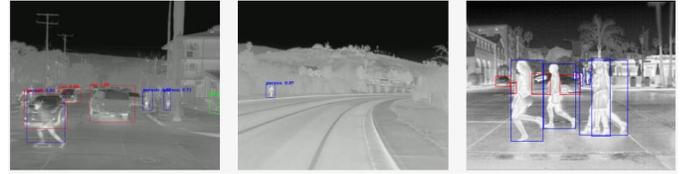

Fig. 3 Examples of Infrared images [15]

Table 2. Summary of evaluation metric for machine tasks

| Task | Evaluation Dataset | Evaluation metric |
|---|---|---|
| Object Detection | OpenImageV6[17] | Mean Average Precision (mAP@0.5) |
| | FLIR [15] | Mean Average Precision (mAP@[0.5:0.95]) |
| | TVD [22] | Mean Average Precision (mAP@0.5) |
| | SFU-HW-object-v1 [23] | Mean Average Precision (mAP@[0.5:0.95]) |
| Object Segmentation | OpenImageV6[17] | Mean Average Precision (mAP@0.5) |
| Object Tracking | HiEve-10[20] | multi-object tracking accuracy (MOTA) |
| Action Recognition | HiEve-10[20] | frame mean average precision (fmAP@avg action) |
| Pose Estimation | HiEve-10[20] | Mean Average Precision (mAP@0.5) |

To evaluate a technology solution for VCM, besides the task performance as measured by the metrics in Table 2, the cost to store or transmit the generated bitstreams is another important factor. The VCM group adopts bits per pixel (BPP or bpp interchangeably) as the measure of bitstream rate (cost) for images used in object detection and object segmentation, computed as follows:

$$BPP = \frac{Total\ bitstream\ size\ in\ bits}{number\ of\ pixels\ in\ the\ souce\ image} \quad (6)$$

Note that we emphasize the source image here because the input image to the task networks can be a scaled version of the source image, which will be further explained in Section IV-E. For videos used in object detection/tracking, action recognition, and pose estimation, bitrate in bits per second is used as the measure of bitstream rate (cost).

We can plot rate-distortion (RD) curves, such as mAP vs. BPP curves to show the performance of proposed solutions. How to quantitively compare the performance of two solutions remains a question. In [25-27] BD-rate is proposed to evaluate the difference between two technical solutions. For example, Fig 4. shows mAP vs. BPP curves for Anchor solution (blue) and Test solution (red). The BD-rate is computed as -14.75%, which shows that the Test solution saves on average about 14.75% bits to achieve the same mAP performance as the Anchor solution.

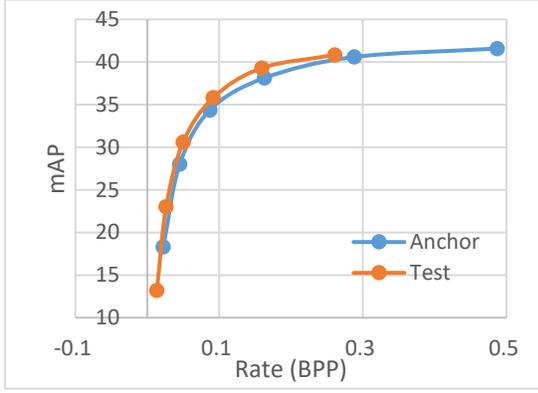

Fig. 4 An example of RD curves for Anchor (blue) solution and Test (red) solution

The VCM group has adopted BD-rate together with its counterpart BD-mAP, BD-MOTA, BD-fmAP as the performance metrics for comparison between the proposed test solution and the anchor solution [6]. We will discuss the anchor solution in Section IV-E.

### D. Evaluation metric for multiple tasks

Another question is how to compare a solution where a bitstream can be used for multiple vision tasks such as the hybrid machine and human vision use case. [25] introduces the concept of weighted precision which combines the effect of machine vision and human vision. Specifically, the weighted distortion $D$ is first computed as follows:

$$D = (1-w)D_m + wD_h \quad (7)$$

where $D_m = 1 - mAP$ is the machine vision distortion metric Note that mAP is used as an example metric and other metrics such as MOTA, fmAP are also applicable; $D_h$ denotes the human vision distortion computed as a weighted normalized mean error (NME) for the three color components Y, Cb, Cr, shown as following formulas:

$$NME_Y = \frac{1}{N_Y(\|Y\|_{max})^2}\sum_{i=0}^{N_Y-1}\|Y'(i)-Y(i)\|^2 \quad (8)$$

$$NME_{Cb} = \frac{1}{N_{Cb}(\|Cb\|_{max})^2}\sum_{i=0}^{N_{Cb}-1}\|Cb'(i)-Cb(i)\|^2 \quad (9)$$

$$NME_{Cr} = \frac{1}{N_{Cr}(\|Cr\|_{max})^2}\sum_{i=0}^{N_{Cr}-1}\|Cr'(i)-Cr(i)\|^2 \quad (10)$$

$$D_h = NME = w_Y * NME_Y + w_{Cb} * NME_{Cb} + w_{Cr} * NME_{Cr} \quad (11)$$

where $N_Y$, $N_{Cb}$, $N_{Cr}$ are number of pixels in the Y, Cb, Cr channels respectively; $\|Y\|_{max}$, $\|Cb\|_{max}$, $\|Cr\|_{max}$ are used to represent the maximum value in Y, Cb, and Cr channels and are often set to be the same value in practice. $(Y(i), Cb(i), Cr(i))$ represents the i-th pixel in the original image or video frame and $(Y'(i), Cb'(i), Cr'(i))$ represents the i-th pixel in the decoded image or video frame; $(w_Y, w_{Cb}, w_{Cr})$ are non-negative weighting factors representing the relative importance of the three color channels and $w_Y + w_{Cb} + w_{Cr} = 1$. Finally, the overall weighted precision wmAP is defined as *1-D,* which can be used to replace mAP to compute the BD-rate.

The VCM group discussed this weighted distortion/precision concept but had difficulty reaching a conclusion on how to decide the weight values. Thus the issue of how to evaluate and compare solutions targeting multiple machine tasks remains.

### E. Anchor Generation

A summary of the anchor generation process is provided in this section. For details of anchor generation, please refer to [6]. First, the anchor generation follows Pipeline 1 in Fig. 2, where the VVC reference software VTM version 8.2 (VTM-8.2) is used to encode the input images/videos and decode the generated bitstreams.

In video coding for human consumption, we often scale input videos to smaller resolutions before video compression and re-scale them back to their original resolution for human consumption to achieve better rate distortion trade-off at certain operating points. Thus, to explore the similar impact of different scales of images/videos on the performance of machine tasks, four scales, i.e., 100%, 75%, 50%, and 25%, of the original images/videos are compressed by VTM-8.2 and decompressed images/videos are fed into the five tasks network as shown in Table 1. An example of such pipeline to generate anchor results is shown in Fig. 5.

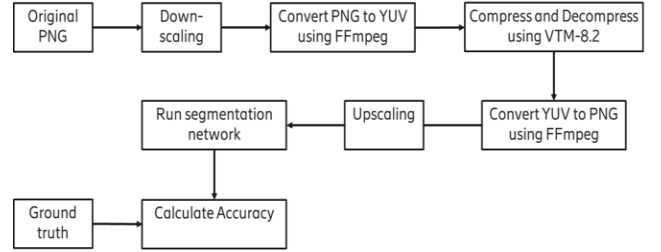

Fig. 5 Pipeline for VCM anchor generation [6]

Note that input images/videos can be in PNG format or other formats such as JPG. The open-source software FFmpeg release 4.2.2 is used to downscale/upscale images/videos and perform format conversion. In addition, for 100% scale, border padding instead of scaling is used at the encoder side to make sure the width and height of an image are even so that the PNG to YUV420 conversion is possible. On the decoder side, the padded border pixels are removed in a 100% scale case. The machine task performance metric is computed using the original images/video resolution such that there is no need to scale the ground truth.

Within the VCM group, there are a lot of discussions on how to utilize the results from the four scales. For example, the results for 100% scale was suggested to be used as VCM anchors for simplicity. On the other hand, to combine the best performance results of the four scales, a Pareto-front curve generated from the RD-curves of the four scales of inputs is suggested to be used as the VCM anchors. An example of the Pareto-front curve is shown in Fig 6.

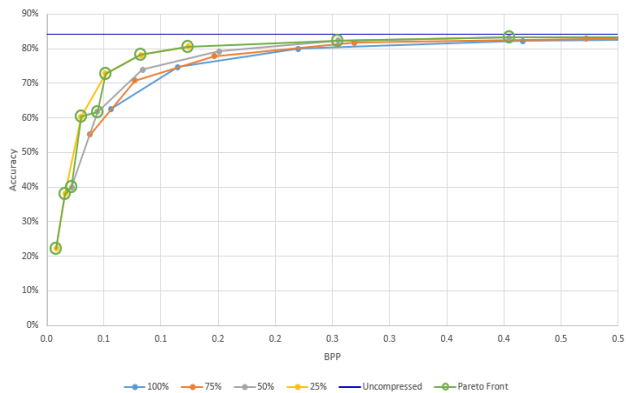

Fig. 6 An example of Pareto-front curves from 4 scales [28]

In Fig. 6, the uncompressed results are obtained by feeding the source images into the task network without the video encoding/decoding process. The green line with circles is the pareto front curve. One thing worth mentioning is that the Pareto-front curve looks zigzag and is not as smooth as the other individual curves. This property sometimes causes unreasonable BD-rate results. In addition, the Pareto-front curve often exhibits a large dynamic range of accuracy. For example, 20% to 85% as shown in Fig 6. The lower part of the dynamic range may not be of interests in practice and a cut-off minimum performance threshold is desirable.

The experts in the VCM group have spent considerable efforts to generate the anchor results for the five machine tasks listed in Table 1, which have been crosschecked by one or two parties to make sure of their correctness. The anchor results are included in [6] and omitted here.

## V. PROPOSED TECHNOLOGIES FOR VCM

Since the creation of the VCM group, different technical solutions related to video coding for machines have been proposed to the MPEG VCM group. For the ease of discussions, these solutions are classified into four categories: (1) Encode output of machine tasks; (2) Encode intermediate feature maps; (3) Learning based image compression; (4) Coding for both human and machine vision. Solutions in these categories will be summarized in the following sections.

### A. Encode Output of Machine Tasks

In this category, the output of machine tasks, such as semantic maps from semantic segmentation, are compressed, where one pixel represents a class label. For example, in [29], the authors utilize the property of the semantic map, i.e., the existence of large single-color areas and object boundaries that can be coded efficiently using chain code or run-length coding, to efficiently compress the semantic map, as shown in Fig. 7(b).

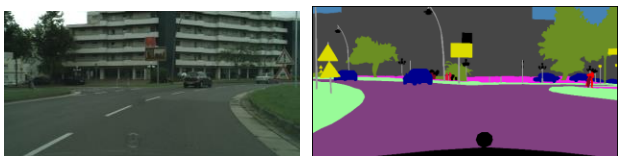

(a) original image        (b) semantic map
Fig. 7 An example of semantic segmentation

Note that this line of work doesn't quite fit into the scope of the VCM group in which videos or features extracted from the videos are compressed and sent through networks to servers for machine vision or hybrid human-machine vision tasks. Additionally, the data volume of the output of machine vision tasks is often quite small and easy to compress, as shown in [30]. Thus, compression of the output of machine tasks is of little interest to the group.

### B. Encode Intermediate Feature Maps

Quite a few technical solutions [31-43] belong to this category, which follows the Processing Pipeline 2 as shown in Fig. 2. Among these solutions, the main difference lies in the following aspects:

#### 1) Different task networks

In early contributions to VCM group, the networks used in feature map compress are quite diverse. For example, Mask R-CNN with Resnet-50 backbone network is used in [31,32]; YOLO v2 is used in [34], etc. In recent meetings, the networks for feature compression tend to converge to those shown Table 1. For example, Faster R-CNN with ResNeXt101-FPN backbone network is used in [37-41] for object detection, and JDE with YOLO3 as the underlying network is used in [43] for object tracking. Using the same networks will enable comparison of different technical solutions.

#### 2) Different partition of task networks

Even with the same networks, different partitions exist in different solutions. For example, for the same object detection task network, i.e., Faster R-CNN with ResNeXt101-FPN backbone network as illustrated in Fig. 7, there are three choices of features maps for compression (or equivalently, three ways of partitioning the network into Part 1 network and Part 2 network):

(1) The output of the stem layer [38-40]

(2) The output of the whole backbone network, i.e., the multi-scale features {P2, P3, P4, P5, P6} [37]

(3) The output of the layer C2 [41]

Different features maps have different data volume. For examples, it is shown in [39] that the data volume of the feature maps from the stem layer is much smaller than the whole multi-scale maps, i.e., {P2, P3, P4, P5, P6}. However, [39] doesn't compare the bitstream size when compression is applied to the two types of feature maps. Therefore, it is not conclusive which feature maps should be chosen from a compression efficiency point of view. In addition, as we know, one reason to compress the intermediate feature maps is to offload part of computation from servers to front-end devices. If features from an early layer of networks are selected for compression, the benefit of computation offloading for servers will be minor. Thus, selection of intermediate layer feature maps needs to take both compression efficiency and computation offloading into account. In addition, the raw data volume of feature maps may not be a good indication of the size of the resulted compressed bitstream since it is closely related how well the compression algorithms can utilize the redundancy in the feature maps.

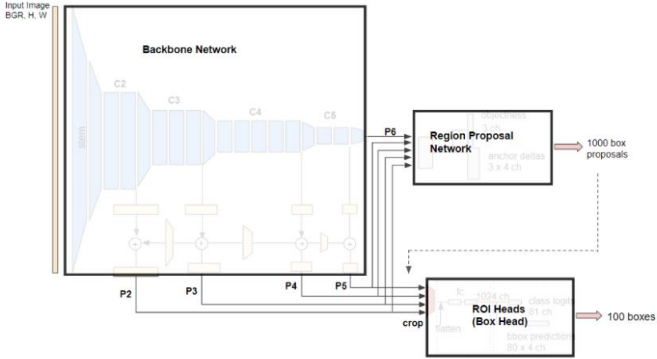

Fig. 7 Architecture of R-CNN FPN [45]

*3) Different Quantization/Normalization schemes*

The output feature maps of the Part 1 network are of 32-bit floating number format, which need to be converted to integer for further compression. For example, uniform quantization is used in [31,32]; To reduce the range of quantized features, a clipping operation is suggested in [32]. In [36], features are normalized using the mean and standard deviation, as shown in the following formula:

$$z = \frac{x-\mu}{\sigma} \quad (1)$$

where $x$ and $z$ denote the feature and the quantized feature; $\mu$ and $\sigma$ denote the mean and standard deviation of the features in the corresponding feature channel. The normalized feature can be uniformly quantized into 8-bit integers using the following formula:

$$\hat{z} = 255 \times \frac{z-z_{min}}{z_{max}-z_{min}} \quad (2)$$

where $z_{max}$ and $z_{min}$ are the maximum and minimum values of the normalized features, respectively. On the other hand, a normalized feature can be further quantized into 2-bits value using the following formula:

$$\hat{z} = \begin{cases} 0, & if\ z < -z_{th} \\ 1, & if\ -z_{th} \leq z < 0 \\ 2, & if\ 0 \leq z < z_{th} \\ 3, & if\ z > z_{th} \end{cases} \quad (3)$$

where $z_{th}$ is the quantization threshold. Simulation results from [36] are illustrated in Table 3.

Table 3. Object segmentation results using Mask R-CNN with Resnet50-FPN backbone network (dataset: Cityscapes and $z_{th} = 1.5$)

| Features | mAP (%) | Data size ratio (features/input) |
|---|---|---|
| Input image | 36.4809 | 1 |
| Original features (32-bit) | 36.4809 | 28.41 |
| Quantized features (8-bit) | 36.4617 | 7.10 |
| Quantized features (2-bit) | 35.4592 | 1.77 |

From Table 1, we observe that the mean average precision (mAP) performance loss from 8-bit quantization is insignificant while the performance loss from 2-bit quantization is around 1%. This indicates much fewer bits than the original 32-bit are required to represent a feature without significant degradation to the performance.

In addition, principal component analysis (PCA) is employed in [34] to reduce the size of the feature maps before applying video compression in a later stage.

*4) Different Feature Packing*

The normalized/quantized features are packed into video frames that are suitable for video encoding. Feature maps of an image can be packed spatially to form one video frame coded by a video codec with intra coding or can be packed temporally to form multiple video frames coded by a video codec with inter coding. Following are a few examples:

(1) Spatial packing: In [35,37], all the multiple scale feature maps {P2, P3, P4, P5, P6} are packed into one feature frame. An example of such packing scheme is shown in Fig. 8(a), where each small block denotes one feature channel of different size. In [38], the 64 feature maps from the stem network are tiled into one feature frame. Specifically, for a given spatial position $(x, y), 0 \leq x < w\ and\ 0 \leq y < h$ where $w$ and $h$ are the width and height of the feature map respectively, the corresponding 64 feature values from the 64 channels form an 8x8 tile. All the 8x8 tiles are packed together to form a feature frame of size $8w \times 8h$ in YUV400 format.

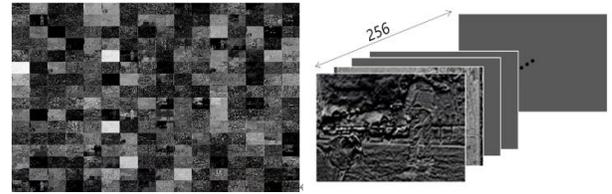

(a) spatial packing    (b) temporal packing
Fig. 8 Example of spatial packing and temporal packing of feature maps [41]

(2) Temporal packing: In [41], the 256 feature maps from the output of C2 layer are treated as a video sequence of 256 fames, as illustrated in Fig. 8(b). It is shown that reordering the feature maps by grouping similar maps together can help improve the coding efficiency.

*5) Different Feature Encoding*

Different encoding methods are used in various solutions. For example, the quantized features are compressed using zip or Huffman coding in [31,32]; VVC codec is used in [34,35,38] while HEVC codec is used in [37]. Note that all the above intermediate feature map compression methods still yield bitstreams much larger than those by directly encoding the source videos/images.

Instead of treating feature maps as regular data or as images or videos, the authors in [42] take the efforts to investigate the statistic property of features maps, i.e., the boundary, histogram and linear correlation characteristics, and propose methods for compression. However, the size of coded bitstreams is not provided in [42]. Hence, we don't know the real benefit of this approach.

In addition, a re-trained deep learning-based image compressor [44] is utilized to compress feature maps data in [40]. Comparable performance as the VCM anchor can be achieved for COCO2017 dataset. This is a new direction and achieves the best results so far.

*6) Summary*

Due to the above differences, it is rather difficult to compare different technology solutions in this category, especially for those early contributions to MPEG-VCM group. With the convergence of machine vision networks, partitioning of networks and evaluation dataset, fair comparison starts to become feasible.
The compression of intermediate feature maps has pros and cons: one benefit of this approach is computation offloading when the computation power in the servers is limited while the edge devices are powerful enough. Moreover, if features generated by the Part 1 network can be used by multiple machine tasks without significant performance degradation, the benefit of computation offloading becomes more obvious [33]. On the other hand,, we observe that the generated bitstreams compressed by traditional data or video compression codec are still much larger than those generated by directly encoding the input images/videos. Deep learning-based approaches show great potential in intermediate feature compression.

*C. Learning-Based Image Compression*

In recent years, deep learning-based image compression becomes an active research area. In [46], the autoencoder framework is utilized to first convert images into latent features which are compressed into bitstreams by the arithmetic encoder with hyper priori; the decoder decodes bitstreams to generate reconstructed latent features, which are further converted to reconstructed images. The network architecture is shown in Fig. 9. In addition, non-local operations are used to capture both local and global correlations among pixels in the original images and latent features; an attention mechanism is adopted to help generate a more compact representation for latent features and hyper priori. Thus, this the method is called Non-local Attention optimization and Improved Context modeling-based image compression (NLAIC for short). The simulation results show that the method can outperform JPEG-2000.

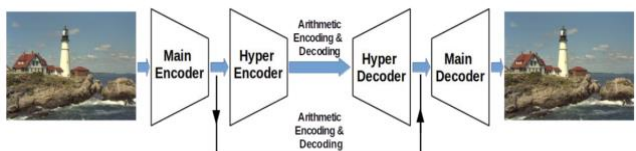

Fig. 9 Overall network architecture [17]

[47] introduces an object-based coding method that utilizes an object segmentation network to partition an image into objects and background using segmentation mask. The objects and background are encoded separately using convolutional neural network as described in [46]. The method achieves similar MS-SSIM performance as that in [46] at low bitrates. In [48] deep learning-based image compression framework, i.e. CompressAI [49], is introduced. This framework includes multiple learning-based image compression algorithms developed in recent years, serving as a good platform for study in this area. Two pre-trained models mbt2018-mean [50] and cheng2020 [51] are selected to evaluate their performance on the object segmentation task against the VVC based anchor solution. From the results shown in Fig. 10, we can observe that the anchor solution is still better than the two models for input images with 100% scale. On the other hand, cheng2020 model has better performance at relatively high bitrates (bit per pixel, or BPP >0.055) than the anchor solution for input images with 50% scale. The baseline curve is generated using original images without compression.

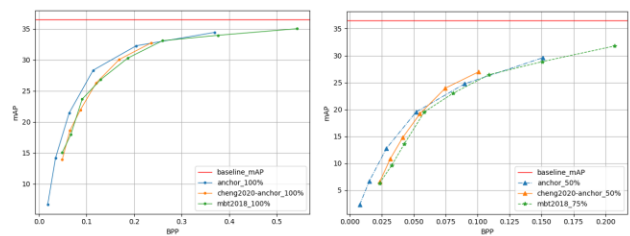

(a) 100% scale    (b) 50% scale
Fig. 10 Object detection performance comparison [48]

One issue for methods in [46-48] is that they are mainly designed for human consumption and it is not clear how the decoded videos/images perform on machine vision or hybrid human/machine vision tasks. Moreover, the compression performance is still not as good as the anchor solution in which the VVC codec is used.

*D. Coding for Human and Machine Vision*

Methods in [52-56] are related to coding for human and machine vision. In [52], deep learning methods are applied to compress the facial image for facial image verification/recognition tasks. Region adaptive pooling strategy is used for adaptive rate allocation and generative adversary network (GAN) is employed as the metric for compression. Semantic distortion together with pixel fidelity and perceptual fidelity metric is utilized when optimizing the network. The simulations show that the method outperform JPEG2000 image codec in terms of PSNR, SSIM and face recognition accuracy at very low bitrates (~0.110 BPP). [53] proposes an object-based image compression approach with semantically structured bitstreams. A decoder can reconstruct the full images or partially reconstructed images with a semantic structure for machine analysis tasks. Simulation results show that the method can outperform BPG for human vision. The object classification performance using full reconstructed images is similar to that using partial bitstreams. Note that [53] is closely related to [5] described in Section II. [54] utilizes the object-based image compression proposed in [47] and demonstrates that the method not only improves subjective visual quality for human vision but also improves the performance for machine vision tasks such as object detection, semantic segmentation and instance segmentation, compared to BPG image codec. Part of the results are shown in Fig. 11. The authors claim one possible reason for the better performance for machine vision tasks is that learning-based image compression may preserve more features that can be exploited by neural networks for machine tasks. Note that since BPG is based on HEVC, it is not clear how the proposed method compares to the VVC codec for both human and machine vision.

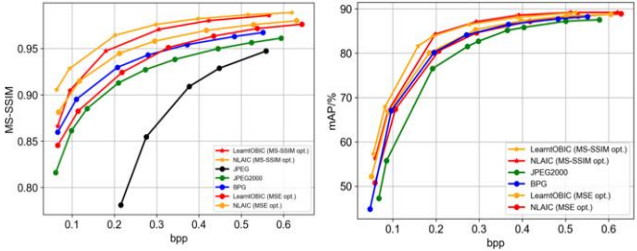

Fig. 11 Performance comparison for human vision and machine vision task, object detection

In [55], the authors propose a method that employs a conditional deep generation network to reconstruct video frames with the guidance of learned sparse points and the key frames coded with the traditional video codec, as shown in Fig. 12. Since only keyframes and sparse key points are coded in the bitstream, higher coding efficiency can be achieved. From the results in Table 4 and Table 5, we observe that the method improves both the video reconstruction quality and also the action recognition accuracy, compared to the HEVC codec. Note that the approach in [55] is also described in [3] as mentioned in Section II.

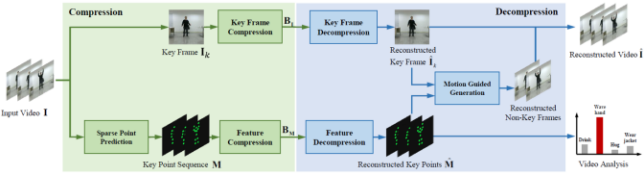

Fig. 12 The pipeline for joint feature and video coding [55]

Table 4 Performance comparison for human vision [55]

| Compression | Bitrate(Kbps) | SSIM |
|---|---|---|
| HEVC | 33.0 | 0.9008 |
| Method in [55] | 32.1 | 0.9071 |

Table 5 Performance comparison for action recognition [55]

| Compression | Bitrate(Kbps) | Accuracy |
|---|---|---|
| HEVC | 16.2 | 65.2% |
| Method in [55] | 5.2 | 74.6% |

In [56], a hybrid rate-distortion optimized video coding method is proposed. This method takes both the fidelity distortion and semantic distortion of the reconstructed video into account. The semantic distortion $D_{MIOU}$ is expressed as follows:

$$D_{MIOU} = -10 \times \ln(MIOU) \quad (4)$$

where MIOU denotes mean intersection over union. $D_{MIOU}$ is modeled as an affine function of the mean square error (MSE) between the source and the reconstructed video based on results from experiments. In the rate distortion optimization (RDO) process, the distortion $D$ is expressed as weighted sum of $D_{MIOU}$ and SSE (sum of square error); the Lagrange multiplier $\lambda$ are expressed as weighted sum of the multiplier for $D_{MIOU}$ and SSE, shown as follows:

$$D = \theta \times SSE + (1-\theta) \times D_{MIOU}$$
$$\lambda = \theta \times \lambda_{SSE} + (1-\theta) \times \lambda_{D_{MIOU}} \quad (5)$$

where $\theta$ is used to trade-off semantic distortion with pixel fidelity distortion. It is demonstrated through experiments that the method improves both performances for human vision (BD-PSNR) and semantic segmentation (BD-MIOU)), shown in Table 6 and 7, where the VVC reference software is used as the anchor and the proposed hybrid RDO based VVC encoder is used as the test solution.

Table 6. BD-MIOU and BD-rate of the reconstructed video

|  | BD-MIOU | BD-rate (%) |
|---|---|---|
| θ =0.75 | 0.0112 | -24.8673 |

Table 7. BD-PSNR and BD-rate of the reconstructed video

|  | BD-PSNR | BD-rate (%) |
|---|---|---|
| θ =0.75 | 0.0316 | -1.0836 |

Note that the hybrid RDO scheme in (5) follows the weighted distortion idea in IV-D to make trade-off between a specific semantic distortion ($D_{MIOU}$) and human vision distortion.

For solutions on coding for human and machine vision, one difficult issue is the modeling and computation of semantic distortion. Different machine vision tasks often require different semantic distortion measurements. Thus, a generic semantic distortion as analogous to PSNR or (MS-)SSIM used in human vision is difficult to obtain but highly desirable.

VI. RESPONSES TO CALL FOR EVIDENCE

In the MPEG meeting in April 2021, the VCM group received five responses [58-62] to the Call for Evidence (CfE) issued in January 2021. We will discuss them and the related proposals in the following sections.

A. End-to-end deep learning-based image compression

An end-to-end codec network is presented in [58]. Cheng2020 network [51] is used for compression while the machine task network is the VCM object detection network. Both networks are joint trained with a loss function taking both object detection loss and MSE error into consideration. During the training the weights for object detection network are fixed and only weights for image compression network are updated. Experiments show that significant performance gain, i.e., -22.80% BD-rate, is achieved for OpenImageV6 dataset. Similar approach is adopted in [63,64] except that the image compression network is different. In [63], a the modified mbt2018-mean network with an inverted bottle-neck architecture encoder is utilized while mbt2018 network is directly used in [64]. As expected, significant performance gain can be achieved in both [63] and [64].

### B. Region adaptive coding

In [59], an image is divided into object region and background region using VCM object detection network. Thus, two images are generated: one contains only foreground object and the other contains only background. Both images are coded using VVC codec with different settings: the background image is coded with high QP while both images are selectively downscaled. At the end, -30.76% BD-rate with pareto-front mAP saving comparing to the FLIR anchor.

### C. Semantically Structed image compression

In [60], a solution with the same framework, i.e., semantically structured image compression, as that in [4] is introduced. The difference is that the networks to extract high- level features and low-level features are simplified and can still achieve the comparable performance. Since the bounding box is generated at the front-end device and sent to the server, the size of the bitstream for bounding boxes is extremely small. Thus, the performance for object detection is far superior to the anchor solution. However, as explained in Section V.A, the group had difficulty to compare this scheme with other object detection approaches. But the concept of the scalable bitstreams is quite interesting to the group and further study is needed.

### D. Improvement of VVC for machine tasks

In [61], the authors investigate the impact of different coding tools in VVC codec on the machine vision tasks. By turning off some combination of coding tools in VVC coding for object detection or object tracking tasks, BD-rate performance gain (-1.6% for object detection) and/or significant encoding complexity reduction (26% for object detection and 43% for object tracking) can be achieved. This indicates that some coding tools in VVC are not optimized for machine tasks, further improvement may be achieved if coding tools in VVC are re-designed or tuned for machine vision tasks. Two other related proposals [65,66] also address the improvement of VVC for machine tasks. In [65], among 24 compression points, i.e., combinations of 6 QP values and 4 scales (100%, 75%, 50%, 25%), the optimal one is selected for each image using a RDO approach for different lambda values. Average BD rate, -29%, can be achieved. In [66], screen content coding tool IBC [67] is turned on to obtain BD rate gain of -1.9%. In addition, as mentioned in V.D, performance improvement for both human and machine vision can be achieved by modifications of the encoder RDO algorithm in VVC reference software [56].

### E. Intermediate feature compression

[62] is the only response addressing feature compression. Instead of using video codec to compress intermediate feature maps, [62] treats intermediate features, i.e., output of the stem layer of Faster R-CNN ResNext101-FPN network, as pure data and employ vector quantization and binary arithmetic coding for compression. Experimental results show that this method is still far inferior to the anchor solution in which images are directly encoded using VVC codec.

### F. Summary

In summary, the two responses [58,59] provide the evidence that object detection can be improved significantly, compared to the anchor solution. On the other hand, no evidences are presented for other machine vision tasks or multi-task cases. That's the reason the VCM group decided to continue soliciting technical evidences. In addition, as indicated in Section VI.D, the performance of video coding for machine using traditional hybrid block-based codec such as VVC can be much improved either through RDO or coding tool selection. This essentially put a higher bar for other type of solutions such as deep-learning approaches to outperform VVC type solutions.

## VII. DISCUSSION AND CONCLUSION

In this paper, we have introduced the recent development activities in the MPEG VCM group for video coding for machine. VCM is an emerging research area targeting the collaborative optimization of video and feature coding for human and/or machine visions. VCM attempts to bridge the gap between feature coding for machine vision and video coding for human vision. As a future direction, there are a number of research areas that need to be explored. For example, is deep-learning-based image/video compression necessary for VCM? Can the intermediate features be compressed more efficiently than compressing the input videos/images directly? How to evaluate a technical solution applicable for multiple vision tasks including hybrid human-machine vision tasks? How to jointly optimize the compression of feature, video, and model stream in a scalable way? Is there a common feature existing for most machine vision tasks? What is the loss function for machine vision system? The answers to these questions will help deepen our understanding of VCM and also help guide the standardization work in the VCM group. This work is expected to call for more evidences of VCM and we encourage more experts from the academia and industry to join the MPEG VCM group and make contributions.